# Navigating Robot Swarms Using Collective Intelligence Learned from Golden Shiner Fish

LIANG HENG and GRACE XINGXIN GAO, University of Illinois at Urbana-Champaign

## 1. INTRODUCTION

Golden shiners are a type of schooling fish that naturally prefer darkness. It had long been believed that individual golden shiners, or at least some of them in a school, could sense the light gradients in the environment, until recently Berdahl et al. [2013] disproved it. In fact, each individual golden shiner does not know where it is dark. Neither does it sense the light gradients in the environment. It only senses the light intensity at its current location, and modulates its speed based on the light intensity. That is, a golden shiner swims fast if it is bright, and slows down if it is dark. However, by moving together as a school, golden shiners always end up in the shady area. This is an example of emergent collective intelligence in nature.

In contrast to how golden shiners swim, navigating networked robot swarms is often a complicated task: it requires knowing where to go [Parker 2000], locating where they are [Filliat and Meyer 2003; Mao et al. 2007], sensing the environment map [Guivant and Nebot 2001; Durrant-Whyte and Bailey 2006], and path-planning based on the destination and barriers in the environment [Qin et al. 2004; Nouyan et al. 2008]. These processes are computationally intensive. Moreover, as the network scales, the computational load and power consumption increase quadratically, or even exponentially.

Inspired by golden shiners, we apply adaptive collective phenomena in biological animal groups to navigating robot swarms. Each individual robot has minimal knowledge of the destination and the environment, as well as minimal communication to other robots. The decision-making algorithm is also as simple as a modification of Gaussian random walk. By swarming together, the collective intelligence navigates the robots to the destination of interest, for example, an information-rich area. The overall computation and power consumption only increases linearly as the network grows. The scale of the network is no longer a computational burden, but a key feature that enables collective intelligence to exceed the sum of its parts. Moreover, the randomness component in individual decisions can help the robot swarm reach the global optimum [Vanderbilt and Louie 1984].

## 2. SYSTEM MODEL

We model a network of agents (robots, golden shiners, etc.) as a simple graph $\mathcal{G} = (V, E)$ [West 2001]. $V = \{1, 2, \ldots, N\}$ is a set of $N$ nodes. $E = \{e_1, e_2, \ldots, e_K\} \subseteq V \times V$ is a set of $K$ edges; an unordered pair $e_k = (i_k, j_k) \in E$ if and only if the distance between the two nodes $i_k$ and $j_k$ is at most the sensing radius $r$. For each node $i$, we define its neighborhood $\mathcal{N}_i = \{j : j \in V, (i, j) \in E\}$.

The nodes' dynamic is based on the gold shiners' movement strategy: each node's movement is a random walk with its speed modulated by the light intensity and its direction affected by its neighbors. Let a complex scalar $p_i$ denote the location of node $i$, we have the following conceptual model:

$$\dot{p}_i = \text{env}(p_i) \, \text{soc}\big(p_i, \{p_j\}_{j \in \mathcal{N}_i}\big). \tag{1}$$

In (1), $\text{env}(p_i)$ represents the environmental factor. For golden shiners, $\text{env}(p_i)$ is a function of the light intensity at location $p_i$. For the scenario of using a swarm of robots to locate the fire source to assist fire fighters, $\text{env}(p_i)$ is a function of heat intensity at location $p_i$.

$\mathrm{soc}(p_i, \{p_j\}_{j \in \mathcal{N}_i})$ represents the social factor. In a golden shiner school, a node senses the location of its neighbors and prefers the swimming direction such that it will stay together with its neighbors. In addition, the node tends to main at least a certain distance to its neighbors in order to avoid collisions.

In addition, both $\mathrm{env}(p_i)$ and $\mathrm{soc}(p_i, \{p_j\}_{j \in \mathcal{N}_i})$ incorporates some randomness. In this abstract, we consider the following modified Gaussian random walk model:

$$p_i[t+1] - p_i[t] = U \exp(\mathrm{j}\, V), \tag{2}$$

where both $U$ and $V$ are real random variables, and $\mathrm{j}$ is the imaginary unit. We assume $U \sim \sigma \chi(2)$, where $\sigma$ specifies the movement speed, and $\chi(2)$ is the standard chi distribution with 2 degrees of freedom.

The movement speed is modulated by the light intensity. If there is a darkest spot at $\rho$, we have

$$\sigma_t = C_1(C_2 + |p_i[t] - \rho|), \tag{3}$$

where the parameters $C_1$ and $C_2$ are positive constants.

The movement direction should ensure that the node move towards the centroid of neighbors but not too close. Thus, we have

$$V = \angle \left( \frac{w}{\mathrm{card}(\mathcal{N}_i[t])} \sum_{j \in \mathcal{N}_i[t]} \flat_s(p_j[t] - p_i[t]) + Z \right), \tag{4}$$

where the function $\angle(z)$ returns the angle of a complex number $z$; and the "hammer" function $\flat_s(z) = (|z| - s) \exp(\mathrm{j} \angle(z))$ reduces the magnitude of a complex number $z$ by $s$ if $|z| \geq s$, or makes the direction of $z$ opposite if $|z| < s$; $\mathrm{card}(S)$ denotes the cardinality, i.e., number of elements, of the set $S$; $Z \sim \mathcal{CN}(0,1)$ is a zero-mean circularly symmetric complex Gaussian random variable, that is, $Z = Z_r + \mathrm{j}\, Z_i$, where $Z_r$ and $Z_i$ i.i.d. $\sim \mathcal{N}(0,1)$; $w \geq 0$ is the weight of the social factor.

## 3. THEORETICAL ANALYSIS

Without considering the environmental factor, each node moves independently without directional preference. The above model reduces to Gaussian random walk with location-modulated speed. This process is a martingale because $E(p_i[t+1]|p_i[t], \ldots, p_i[1], p_i[0]) = p_i[t]$ [Bass 2013].

Let us consider a simplified one-dimensional scenario. Since all nodes move independently, in the following analysis we drop the subscript $i$. For simplicity in notations, we use $x_t$ to denote the node's location at time $t$. Without loss of generality, we assume the darkest spot at $\rho = 0$. The dynamic model is thus given by

$$x_{t+1} \sim \mathcal{N}(x_t, C_1^2(C_2 + |x_t|)^2). \tag{5}$$

Let $g_\mu(x)$ denote the probability density function (pdf) of $\mathcal{N}(\mu, C_1^2(C_2 + |\mu|)^2)$. Clearly, the node's location at time $t = 1$ has the following pdf:

$$f_{x_1}(z) = g_{x_0}(z). \tag{6}$$

The node's location at time $t = 2$ has the following pdf:

$$f_{x_2}(z) = \int_{-\infty}^{\infty} g_{x_0}(x_1) g_{x_1}(z) \,\mathrm{d}x_1. \tag{7}$$

Similarly, the node's location at time $t$ has the following pdf:

$$f_{x_t}(z) = \int_{-\infty}^{\infty} \cdots \int_{-\infty}^{\infty} g_{x_0}(x_1) g_{x_1}(x_2) g_{x_2}(x_3) \cdots g_{x_{t-2}}(x_{t-2}) g_{x_{t-1}}(z) \,\mathrm{d}x_1 \cdots \mathrm{d}x_{t-1}. \tag{8}$$

Unfortunately, there is no closed-form expression of $f_{x_t}(z)$ when $t \geq 2$.

Fig.1 shows $f_{x_t}(z)$ for $t = 1, 2$, and $3$ computed using numerical integration. With a very high probability, the node quickly moves to the darkest spot, although the movement in nature is a memoryless random walk without directional preference.

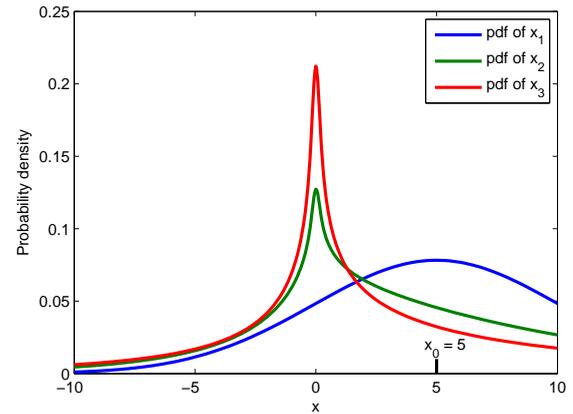

Fig. 1. Probability density function (pdf) of node's location $f_{x_t}(z)$ for $t = 1, 2$, and $3$ under the initial condition $x_0 = 5$ with the parameters $C_1 = 1$ and $C_2 = 0.1$.

## 4. SIMULATION RESULTS

Fig. 2 shows simulated robot swarm behavior based on the navigation strategy of golden shiner fish. Fig. 2 (a) shows the initial distribution of 100 robot nodes, randomly and uniformly distributed over the square area $[-0.5, 0.5]^2$. The darkest spot is at the center. Fig. 2 (b) shows the distribution after 100 steps with only social factor and no environment factor. The individual nodes make decisions based on other nodes' locations and movements. They do not move according to the light intensity cues. The nodes are still quite spread out, although some nodes have already started to cluster at the center. Fig. 2 (c) shows the distribution after 35 steps with both environment and social factor. In addition to speeding up or down based on the light intensity, the individual nodes also tend to follow other nodes' movements and stay with the swarm. The phenomenon of three clusters of nodes near the darkest point is observed. Fig. 2 (d) shows the distribution after 70 steps with both environment and social factor. All nodes have navigated to the darkest area. The results demonstrates that it is feasible and effective to navigate swarms with the simple, highly distributed algorithm learned from golden shiners. The environmental and social factors have complementary benefits: the former enables the nodes to coverage to the right target, while the latter expedites the convergence.

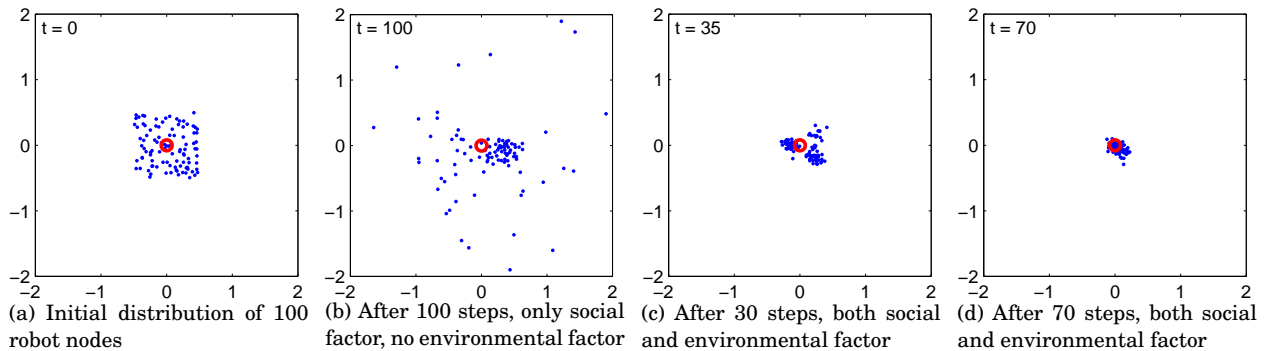

(a) Initial distribution of 100 robot nodes  (b) After 100 steps, only social factor, no environmental factor  (c) After 30 steps, both social and environmental factor  (d) After 70 steps, both social and environmental factor

Fig. 2. Simulated robot swarm behavior based on navigation strategy of golden shiners (darkest spot at the center, $C_1 = C_2 = 0.1$, $r = 0.2$, $w = 20$, $s = 0.08$).

## 5. CONCLUSION

We have proposed a method inspired by the collective intelligence of golden shiner fish to navigate robot swarms. The theoretical analysis and simulation results show that our method 1) promises to navigate a robot swarm with little situational knowledge, 2) simplifies control and decision-making for each individual robot, 3) requires minimal or even no information exchange within the swarm, and 4) is highly distributed, adaptive, and robust.